# Operator Selection While Planning Under Uncertainty

**Todd Michael Mansell**[*]
Department of Computer Science
The University of Melbourne
Parkville, Victoria 3052, Australia.
email: tmm@cs.mu.oz.au.

**Grahame Smith**
Cooperative Research Centre for Intelligent Decision Systems
723 Swanston St,
Carlton, Victoria 3053, Australia.
email: gsmith@cs.mu.oz.au

## Abstract

This paper describes the best first search strategy used by U-Plan (Mansell 1993a), a planning system that constructs quantitatively ranked plans given an incomplete description of an uncertain environment. U-Plan uses uncertain and incomplete evidence describing the environment, characterises it using a Dempster-Shafer interval, and generates a set of possible world states. Plan construction takes place in an abstraction hierarchy where strategic decisions are made before tactical decisions. Search through this abstraction hierarchy is guided by a quantitative measure (expected fulfilment) based on decision theory. The search strategy is best first with the provision to update expected fulfilments and review previous decisions in the light of planning developments. U-Plan generates multiple plans for multiple possible worlds, and attempts to use existing plans for new world situations. A super-plan is then constructed, based on merging the set of plans and appropriately timed knowledge acquisition operators, which are used to decide between plan alternatives during plan execution.

## 1 INTRODUCTION TO U-PLAN

Traditional planning systems describe the planning problem as the composing of a course of action that transforms the world from a given initial state to a desired goal state. This description makes two assumptions about the planning domain: complete and accurate information about the world is available; and the environment is static. These assumptions ensure that a constructed plan can be successfully executed in such an ideal world. However, such planners rarely produce plans that work in the real world, that by its nature is dynamic and its description is imprecise. To devise a useful plan given uncertain and/or incomplete information about a dynamic environment requires a planner to avoid these assumptions.

A major difficulty when planning given incomplete and uncertain information about the environment is that it is not possible to construct one initial state that precisely and unambiguously represents the world. U-Plan uses a possible worlds representation, where the available initial information is used to construct every possible initial state of the world. Associated with each possible world is a numerical measure of belief specifying the degree to which the evidence supports each possible world as the one that represents the true state of the world.

U-Plan does not construct a state-based search tree, but constructs a strategy hierarchy which is a decision tree like structure, where the nodes in the hierarchy represent a continuous transition of actions from the strategic (at the root node) to the tactical (at the leaf nodes). The strategy hierarchy can be represented as an AND/OR search tree, the root node representing the strategic goal of the system, and the leaf nodes representing the tactical details of how the goal is to be achieved. Each node in the tree is a subgoal representing the current goal and world, and certain pairs of nodes are connected by arcs representing the application of a reduction operator that produces each subgoal.

U-Plan utilises a set of (predefined) goal reduction operators that encode how a planning goal is reduced by the operator's application. What results is a planning hierarchy tree where the goals are broken up into subgoals by the goal reduction operators. This allows us to first make the strategic decisions, which then guides all other decisions down to the tactical implementation of the subgoals. A measure of expected fulfilment

---

[*] Present address Materials Research Laboratory, PO Box 50, Ascot Vale, 3032, Melbourne, Australia.

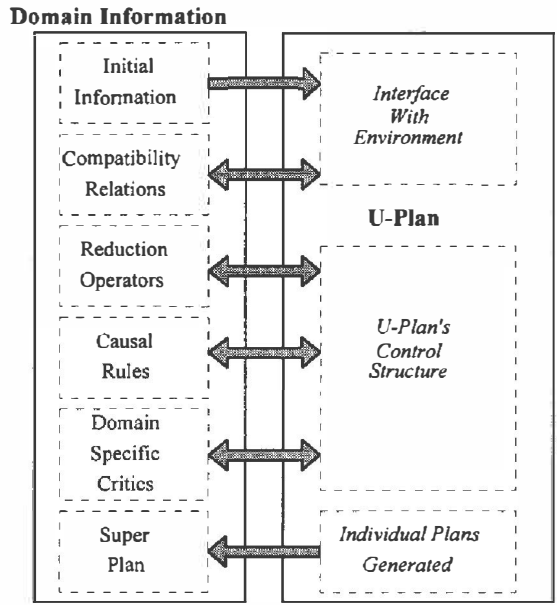

Figure 1: The interaction of U-Plan with domain specific information.

(section 4) is used when selecting which operator to apply next.

In support of hierarchical planning, each possible world is described at a number of predefined abstraction levels, resulting in decisions being made based on a description of the world at a suitable level of detail.

U-Plan constructs a plan for one possible world at a time, the first plan being constructed for the possible world with the greatest likelihood of representing the true world. Before a plan is constructed for the next possible world, the suitability of reapplying an existing plan to this world is assessed. Associated with each plan is the set of possible worlds it works for. If a plan partially works for another possible world (e.g. the strategy works but some of the detail is different), then that part of the plan is used for this possible world, and planning continues from where the plan failed. When a plan exists for every possible world, the operator order of all the plans is combined to obtain a single planning tree that branches when the operator execution order differs. At this point the ability to acquire additional knowledge is used. At each branch, a knowledge acquisition operator can be inserted to gather current information about the state of the world and so determine which action in the planning tree to carry out next.

This planning algorithm has been tested on an air combat domain. In this example, the goal is to successfully attack an aggressor aircraft, given only partial information about his location, type, and status. A number of strategies exist on how the attack should be carried out. Each strategy uses a different method of attack, and therefore has a different probability of success, and a different degree to which it fulfils the goal.

The planner, U-Plan, has been coded in Lisp on a Symbolics 3600 series machine. This planning system is not a single domain planner, the code used to characterise the air combat domain is separate from the code used to represent the body of U-Plan. Figure 1 demonstrated the domain specific initial information required by U-Plan to carry out planning. The initial information describing the state of the world and compatibility relations (Lowrance et al 1991) are passed to that part of U-Plan that interfaces with the environment; while the domain specific critics, causal rules, and the reduction operators are used by U-Plan's main control structure for plan generation.

The specific domain types intended for use by U-Plan are referred to in this paper as *emergency response* problems. That is, domains in which action would be required before complete and precise information could be obtained, but where a general course of action that achieved the goals was produced. These *emergency response* problems should include the circumstances where a number of planning solution were likely to exist for each possible world. Hence, one of the planning objective will be to generate a plan for one possible world, that could be applied, in part or whole, to a number of other possible worlds.

A number of alternative plans are likely to be needed to cover the sets of possible worlds that describe the environment. The final plan produced by U-Plan should include a method for choosing between these alternatives, in the form of a measure of the evidence attributed to the plan branches. The final output should also identify action sequences common to a group of alternative plans, that can be applied before one must choice to acquire additional information, or select between alternative plans using a measure of belief.

Central to planning using U-Plan is: the set of states are represented at several abstraction levels; the selection of reduction operators is not based on the modal truth criterion, but dependent on a calculation of expected fulfilment; the system will plan to acquire additional knowledge when it is advantageous to do so, and an attempt is made to apply an existing plan to more than one initial state.

## 1.1 THE AIR COMBAT DOMAIN

The air combat domain is dynamic and requires agents to act given uncertain and incomplete information. To generate a plan for aircraft operating in such a domain requires the consideration of a large number of plan strategies. The role of an AI planner in such a domain is to construct a plan that best fulfils the stated goals using only available information, with the option to plan to acquire additional information if required.

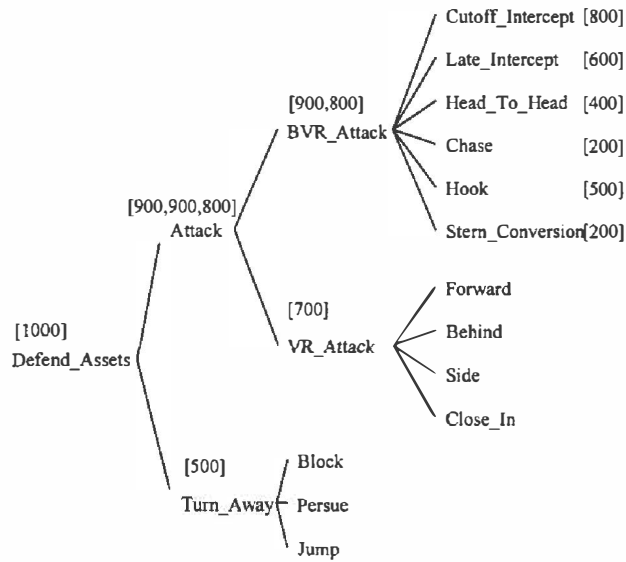

Figure 2: The strategic portion of the abstraction hierarchy for the simplified air combat example.

In our implementation of this domain (the abstract portion of which is given in figure 2), we consider two agents: the *defender* (whose objective is to defend himself and a designated airspace) and the *aggressor* (who is invading the airspace controlled by the defender). While the various actions available to the agents in this domain are diverse, certain patterns in the more abstract strategies can be identified. For example some of the actions available to the defender are *target monitoring, strategy selection, strategy implementation*, and *evaluation of attack strategy*. To operate in such an environment requires a sophisticated planning and reasoning system. The strategies available to U-Plan are centred around the selection of either a Beyond-Visual-Range Attack or a Visual-Range Attack. The different ways such a strategies can be carried out are numerous, each one involves a unique course of action.

It is intended that U-Plan generate a suitable course of action for the defender aircraft given only the information available to the aircraft at the time. This plan should intend to acquire necessary information when appropriate, have a high likelihood to success, be applicable to as many possible worlds as feasible, and work for the worlds that are most likely to be true. The plan produced by U-Plan is intended for use in post mission analysis of the defender aircraft, not as a real time planning aid (as the system is not running in real time).

## 2 STATE REPRESENTATION

When an incomplete model of the world is all that is available, a set of initial states can be used to describe the alternative environments. U-Plan employs a set of initial possible states (P-states) to describe what might be true of the world. A P-state, ps(a), is a complete description of one possible world using propositional statements. Each P-state is described hierarchically with n levels of abstraction, $(ps(a)=\{\ell_1(a) ... \ell_n(a)\})$ where n is domain dependent and selected during knowledge engineering. The level $\ell_i(a)$ is a complete description of a world at the $i^{th}$ level. The highest level of abstraction gives a coarse description of the state of the world. The lowest level gives a detailed view of the world. Intermediate levels provide the description required to make a smooth transition between both extremes. Compatibility relations (supplied at knowledge engineering as shown in figure 1) are used to construct and maintain consistent representations of the world at all levels.

Associated with each P-state is a two valued quantitative measure (an evidential interval (Shafer 1976, Lowrance *et al* 1991)) that characterises the weight of evidence that supports the P-state accurately describes the true state of the world. This information is used by U-Plan in determining the order in which P-states are planned for, and the final execution order of operators. A detailed description of how P-states are generated and used can be found in (Mansell 1993a).

## 3 REDUCTION OPERATORS

Planning operators represent actions that the system may perform in the given domain. The role of an action is to change the state of the world, the aim of an operator is to represent how applying that action will change the system's view of the state of the world. U-Plan uses reduction operators to give alternative methods for achieving the goal at a lower level of abstraction, or at the tactical level it describes the direct effects of an action on the P-state. These are SIPE-like operators (Wilkins 1988) where the closed world assumption is implemented, and hierarchical planning used.

Each operator contains information about how and when it is to be applied to the P-state (Mansell 1993a), using the following description.

**Name:** Name of the operator.

**Abstraction Level:** The P-state's level of abstraction used by the operator for selection and implementation.

**Necessary Preconditions:** The necessary preconditions slot represent what must be true of the world before the operator can be applied. This slot includes propositions that must be true of the world but are unable to be altered, or are prohibitively expensive to do so. Alternatively, propositions that represent aspects of the environment that are to be observed and not changed are included.

**Satisfiable Preconditions:** Satisfiable preconditions designate propositions that must be made true of the world before the action can be applied. This is usually pieces of tactical detail that should be true of the world before application of the operator.

**Plot:** Step-by-step instructions on how to perform the action represented by the operator. This includes a description of the goal reduction operators that are applied at the next level of abstraction, and it's fulfilment, measuring the degree to which it achieves the goal of the present operator. Or, at the lowest level of abstraction, how the operator changes the P-state.

**Probability:** A function for calculating the probability of the reduction operator succeeding given the current P-state is also included. The availability of such a function is domain specific and not always easily obtained. In the air combat example discussed here the function is obtained empirically (based on historical data). The probability of success does not provide sufficient information to select a reduction operator as it does not take into account the goals of the system. It is for this reason that associated with each reduction operator is a measure of fulfilment.

**Postconditions:** What the operator achieves.

**Planfail:** What to do if the operator fails during planning.

U-Plan uses a deductive causal theory (Wilkins 1988) to deduce the context dependent effects of applying a reduction operator to a P-state. The effects that are deduced are considered to be side effects, where those that are introduced directly by the reduction operator are the direct effects. The use of deduced effects simplifies the description of the operators by removing the need for extensive add and delete lists. After the application of each reduction operator a set of triggers are used to determine if the world has been changed in such a way that the deductive rules need be applied. If so, the deductive causal theory is used to change the P-state to be consistent with all the effects of an action. The side effects of applying any reduction operator are recorded in the planning hierarchy tree. These rules are domain specific and supplied at knowledge engineering (fig. 1).

# 4 OPERATOR SELECTION

Many classical planning systems use a state-based search strategy to solve planning problems. To find a solution one applies operators to a state description until an expression describing the goal state is found. U-Plan uses a quantitative measure, called expected fulfilment, in an abstraction hierarchy to guide the selection of operators.

A plan constitutes the successive application of reduction operator from the highest level of abstraction (i.e. the goal function) down to the most tactical detail. In the air combat example (fig. 2) the goal of defending specified assets is accomplished by first choosing whether to attack or turn away the aggressor, through specific strategies and manoeuvres available, down to the detailed implementation of a specific manoeuvre.

The following sections outline how the reduction operators are selected and implemented, and the process that is continually reviewing these decisions.

## 4.1 EXPECTED FULFILMENT

Goals in many domains dealing with uncertainty (for instance the air combat domain) are not precise requirements. Many general goals can be fulfilled to various degrees by achieving alternative subgoals. However, not all subgoals are equally likely to be achieved. We adopt an approach to planning by determining a course of action that is likely to maximise the expected fulfilment of our goal. Consequently, our plans are not exhaustive. They do not elaborate all the alternative actions required in all possible worlds. Rather, they specify alternative actions that are likely to maximise the expected fulfilment of our goal in the possible worlds that are consistent with our partial description of the world.

Expected fulfilment is a quantitative measure used to rank the reduction operators which achieve the goals of the active operator (i.e., the next reduction operator chosen to be expanded) for selection purposes. For example, if *attack* is the active operator who's goals we wish to achieve, the expected fulfilment for *BVR Attack* and *VR Attack* (the operators that achieve this subgoal) is calculated and used as a basis for the selection.

Probability theory provides an effective method for choosing actions capable of producing consistently accurate choices. Information such as intelligence reports, preferences, and raw data can be encoded and manipulated by a probabilistic inference engine to produce useful recommendations. Whereas probabilities are used to represent the likelihood of events, fulfilments are used as a local measure of the degree to which the consequent of the action achieves the intended goal and the desirability of that action. The term fulfilment is used to capture both the essence of utility scaled according to the desire to use a particular approach. The term utility is not used in this description to avoid confusion, as there are subtle differences in how they are used and what they represent.

The expected fulfilment is used as a measure of an action's likelihood to produce the consequent that achieves the agent's goals. If we use the measure $F(c)$ to represent the degree of fulfilment of consequent $c$, then the overall expected fulfilment associated with action $a$ is given by:

$$EF(a) = F(c)P(c|a, ps), \qquad (1)$$

where, $P(c|a,ps)$ is the probability of achieving consequence $c$, conditioned upon selecting action $a$ and observing evidence contained in the P-state, $ps$.

For example, to calculate the expected fulfilment of the *Attack* operator in figure 2, the probability of successfully executing an attack in the given P-state is multiplied by the degree of fulfilment obtained by executing the action (depicted in figure 2 as the first number shown in square brackets above the operator).

The expected fulfilment of action $a$ $EF(a)$ is regarded as a gauge of the merit of action $a$. The expected fulfilment is used as a procedure for choosing among alternative (or competing) actions. When given the choice between two action (eg, *Attack* and *Turn-Away*) the selection is based on the action that yields the highest expected fulfilment (ie, $EF(Attack)$ or $EF(Turn-Away)$). This result will depend on the description of the P-state when the selection is made. This process can be thought of as establishing a rank order in which one should attempted to apply a set of actions.

### 4.2 APPLYING AN OPERATOR

Once the reduction operators that achieve the goals of the parent operator have been ranked using the expected fulfilment calculation, they can be tested to determine their suitability to the P-state. The successive application of desirable operators to the given P-state then takes place in order of greatest expected fulfilment. If the necessary preconditions of a reduction operator are true in the active P-state, then the reduction operator is provisionally selected, else the plan fail for that operator is applied (this usually involves backtracking). When a reduction operator that satisfies the necessary preconditions has been found, the satisfiable preconditions are tested. If any of these are not true, U-Plan can attempt to satisfy them using reduction operators of equal or lower abstraction. If these preconditions are not satisfied, the operator is rejected, and it's planfail procedure is implemented.

Once both sets of preconditions of the reduction operator can be shown to be true in the active P-state, the operator is accepted and its plot can be applied. The plot represents the effects the reduction operator has on the state of the world, and the subgoals that may be used to achieve this goal. When applying the plot, the next level of the strategy hierarchy is exposed, and again the subgoal with the highest expected fulfilment is selected to be expanded next. The plot of operators represents actions at the lowest level of abstraction specify how the P-state is physically changed by their application.

This process of applying operators continues until the next layer of the strategy hierarchy has been exposed. At this point, the earlier selection of specific actions are reviewed as described below.

### 4.3 REVIEWING SELECTED OPERATORS

When constructing a strategy hierarchy it is possible that as a plan's detail is filled out it becomes less likely to succeed. This is partially because the initial strategic decisions are based on information at a more coarse level of abstraction. As the plan is expanded and more tactical decisions about the implementation of specific strategies are made, the expected fulfilment of specific plan branches may decrease. The other reason this occurs is that when the expected fulfilment is calculated for an operator, the fulfilment component is an optimistic assessment of that actions ability to achieve the goal of its parent. However, as planning continues, it is likely at a lower level of abstraction, that a less than ideal actions may have to be applied, therefore reducing the true expected fulfilment of the plan branch.

This makes it important to review earlier decisions while planning. After the application of a group of reduction operators, U-Plan compares the expected fulfilment of the current subgoals, with those of previous subgoals, and determines if they fall below the previous values, less an offset. Including an offset is an iterative deepening strategy[1]. The offset value will depend on the difference in abstraction level of the subgoals. It is expected that as the system uses lower level information the expected fulfilment of the plan will decrease. This offset value helps avoid the problem of the system jumping around from branch to branch in the strategy hierarchy.

Previous operator selections are reviewed whenever a new layer in the abstraction hierarchy has been generated. This process is used to ensure the decision to take a certain planning direction is still favourable. In the air combat domain this means that as planning continues down to the lowest level, the decision to use an *Attack*

---

[1] A number of iterative deepening strategies exist that can be applied to this task of selecting a suitable offset between different abstraction levels.

instead of a *Turn_Away* may be reviewed a number of times throughout plan generation.

To review these selections one must have a way of *updating* the EF values calculated for these nodes in the abstraction hierarchy based on the detail added to that nodes planning branch. A set of update rules are used to re-evaluate the fulfilment and probability of each operator expanded given the most recent planning developments. The update rules used depend on whether the node produces an AND or OR branch in the tree. Referred to as an AND node expansion or an OR node expansion respectively in this paper.

In the case when an operator is expanded and produces an OR branch in the abstraction hierarchy, the update rules used to determine the fulfilment and probability of a *parent* node given a set of possible *children* are given by:

**Rule 1:** When the application of a reduction operator results in an OR node expansion (as defined above) the probability of action, *parent*, can be updated using the following rule:

$$P(parent) = \{P(child) \Big|_{children}^{MAX} EF(child)\}, \quad (2)$$

where *children* is the set of child operators that have been expanded out and may be successfully applied to the current P-state. Simply stated, rules 3 and 2 tell us that the fulfilment and probability values for a parent of an OR node is equal to the fulfilment and probability of the child node with the greatest expected fulfilment that has not been ruled inapplicable to the P-state (i.e. due to failure of preconditions during expansion).

This states the updated probability of the *parent* is equal to the probability of the child action with the greatest expected fulfilment at the next level of abstraction.

**Rule 2:** When the application of a reduction operator results in an OR node expansion (as defined above) the fulfilment of action, *parent*, can be updated using the following rule:

$$F(parent) = \{F(child) \Big|_{children}^{MAX} EF(child)\}. \quad (3)$$

This states the updated fulfilment value of the *parent* action is equal to the fulfilment of the child action with the greatest expected fulfilment.

The justification for the update of an OR node expansion is straight forward. The goals of the parent can only be achieved by the application of an operator in the plot, and the agent may only execute one of these actions in the final plan. Hence, the reduction operator selected to achieve the goals of the parent represents the actions of the parent operator at the next level of abstraction. It is for this reason the probability and fulfilment of the parent operator are set equal to the probability and fulfilment (respectively) values of the child operator included in the plan. The selection of a child operator to be included in the plan is based on choosing the child operator, from the set of child operators, with the greatest weighted EF value.

The updating of the parent reduction operator at an AND node involves updating the probability and fulfilment as follows:

**Rule 3:** When the application of an operator, *parent*, results in an AND node expansion (as defined above) the probability of action, *parent*, can be updated by setting it equal to the product of the probabilities of its children actions at the next level of abstraction. That is:

$$P(parent) = \prod_{children} P(child). \quad (4)$$

As U-Plan produces totally ordered plans, a conditional relationship between the actions in an AND expansion may seem appropriate. However, such constraints are on the information flow between the reductions operators common to the AND expansion. The prerequisite for the expansion of a reduction operator is that its preconditions must be met. The argument for including a conditional relationship on the ordering of reduction operators stems from the dependence some actions may have on the successful execution of a previous action. That is, the preconditions of a reduction operator may require the P-state be altered by anterior reduction operator. However, the application of a specific reduction operator in this circumstance is secondary to the effect that operator has on the P-state. Hence, when updating the parent of an AND expansion, the probabilities of the children operators are considered to be conditional solely on the P-state.

The actions in an AND expansion are assumed to be independent of each other given the constraints outlined in the preconditions of each child reduction operator; and conditional on the P-state they are applied to. The updated probability of a parent action, $A$, given the children actions $\{a_1, a_2, ..., a_n\}$ are assumed to be independent, and are given by:

$$P(A|ps) = P(a_1|ps).P(a_2|ps)...P(a_n|ps). \quad (5)$$

Hence, the probability of the parent of an AND node expansion is equal to the product of the children given the P-state.

**Rule 4:** When the application of an operator *parent* results in an AND node expansion, (as defined above) the fulfilment of action, *parent*, can be updated as being equal to the fulfilment of the child action with the lowest fulfilment at the next level of abstraction. That is:

$$F(parent) = \{F(child) \Big|_{children}^{MIN} F(child)\}. \quad (6)$$

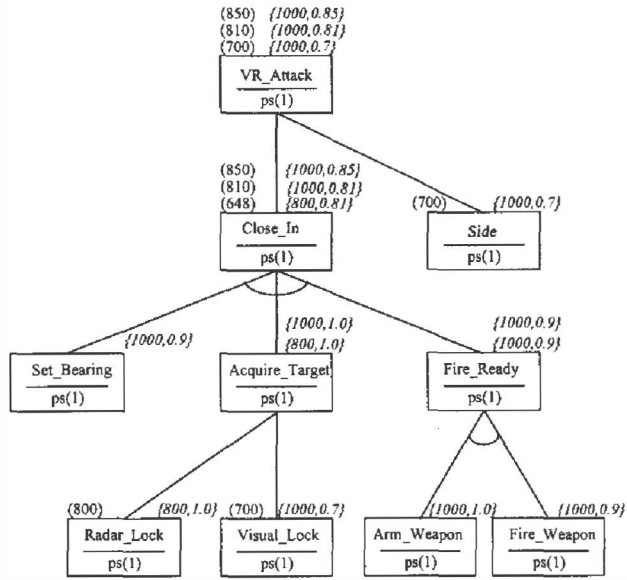

Figure 3: An example of the abstraction hierarchy for a simple *Close_In* manoeuvre operator scenario that demonstrates the updating of fulfilments and probabilities. Progressive (EF) and *{fulfilment, probability}* values are given for each operator.

The justification for updating fulfilments in this way is based on the original distribution of fulfilment and the attitude toward risk. In an AND node expansion, the fulfilment of a *parent* operator is transferred to the children. It is assumed that the goals of the *parent* operator can only be achieved by the application of the specified set of reduction operators (the *children*). As the operators are assumed to be independent and equally important in the eventual success or failure of the *parent* operator, they are each assigned a fulfilment value equal to that of the *parent* operator, to maintain the integrity of the fulfilment process. If, at some later stage in the plan, one of the *children* of an AND node expansion is assigned a lower fulfilment value, this information should be included in the update of the *parent* operator. The reason for this is, as the operators in an AND node expansion are assumed independent and equally important, reducing the desirability of one of the required must reduce the desirability of the *parent* operator. A number of approaches to this problem could have been taken. U-Plan assigns the *parent* operator the lowest of all the *children*'s fulfilments, there by allowing U-Plan to consider the maximum effect on the plan. Some may consider this as taking the pessimistic alternative, as this is assuming the worst. However, the intention is to offer a cautious course to be taken.

In the subset of the air combat example given in figure 3, a simple scenario for a *Close_In* manoeuvre operator is evaluated. The expected fulfilment for this operator is calculated (the first number in the parenthesis on the top left of the operator node, i.e., 850) based on the fulfilment and probability values (shown in the braces on the top right of the operator node, i.e., 1000 and 0.85 respectively) obtained from the operator. On expanding the *Close_In* operator, the next level of the plan is uncovered. This shows that the *Set_Bearing*, *Acquire_Target* and *Fire_Ready* operator are to be applied. The fulfilments and probabilities for these are calculated and shown in braces below the operators. As this is an AND operation, update rules 3 and 4 are used to update the probabilities and fulfilments for the parent, *Close_In*, operator (shown in the second set of braces above the operator, *{1000,0.81}*). These updated values are used to calculate the updated EF value for the *Close_In* operator (i.e., (810)).

Figure 3 also includes an example where rules 1 and 2 are used to update the probabilities and fulfilments for the parent of an OR node. In this case the *Acquire_Target* can be achieved by either a *Visual_Lock* or a *Radar_Lock*. The *Visual_Lock* is chosen as it has the higher EF of the two. However, the *Visual_Lock* has a lower fulfilment than its parent that is propagated back through the branch. It should be noted at this point that, as a result of propagating these values back up the branch, the *side* operator becomes favourable over the expanded *Close_In* branch.

### 4.4 SENSITIVITY OF EXPECTED FULFILMENTS

A major advantage of using U-Plan when planning in an uncertain and incomplete domain is it produces a super-plan, that is a merging of a small (relative to the number of possible worlds) number of plans (Mansell 1993a). This super-plan incorporates the intention to acquire the additional information required to allow the smooth execution of operators. In most cases, U-Plan produces a super-plan in less time than it would take a traditional planner to produce one plan for every possible world (see (Mansell 1993a, Mansell 1993b) for more details on these results).

As U-Plan uses the EF measure to select between alternative reduction operators, it is important to know the certainty with which one knows the expected fulfilment for each action. However, the probability and fulfilment values are not always known precisely. For this reason, choosing between two actions, the possible error in the EF of the actions can be represented in the form of an error range on the EF. If the EF for two of the actions have an error range that overlaps, the action selected may not have the greatest expected fulfilment. It is therefore important to have the facility to work out these errors bounds and determine if the EF ranges overlap.

To understand the accuracy of probability and fulfilment information required by competing reduction operators, let us consider a two operator example. Given two reduction operators, $a$ and $b$, with probabilities, $p_a$ and $p_b$, with possible errors $\alpha p_a$ and $\gamma p_b$. These reduction operators also have fulfilments, $f_a$ and $f_b$, with their possible errors, $\beta f_a$ and $\delta f_b$. The expected fulfilments, $EF_a$ and $EF_b$, for these actions can be calculated as follows.

$$\hat{p}_a = p_a \pm \alpha p_a \qquad \hat{p}_b = p_b \pm \gamma p_b$$
$$\hat{f}_a = f_a \pm \beta f_a \qquad \hat{f}_b = f_b \pm \delta f_b$$

$$\hat{EF}_a = \hat{p}_a \cdot \hat{f}_a \qquad \hat{EF}_b = \hat{p}_b \cdot \hat{f}_b$$
$$= p_a \cdot f_a + \varepsilon_a \qquad = p_b \cdot f_b + \varepsilon_b$$

where
$$\varepsilon_a = \pm \alpha p_a f_a \pm \beta p_a f_a \pm \alpha \beta p_a f_a,$$
$$\varepsilon_b = \pm \gamma p_b f_b \pm \delta p_b f_b \pm \gamma \delta p_b f_b.$$

Given that $EF_a > EF_b$, $EF_a$ must differ from $EF_b$ by the maximum possible error, which requires:

$$EF_a - EF_b \geq \left|\text{Max }\varepsilon_b^+\right| + \left|\text{Max }\varepsilon_a^-\right|$$

Viewing this diagrammatically,

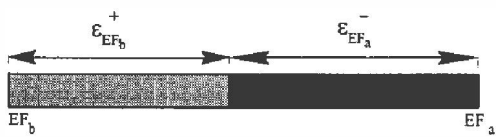

Assuming $\left|\text{Max }\varepsilon_a^+\right| = \left|\text{Max }\varepsilon_b^-\right|$, that is, the total degree of error in the expected fulfilments is equally distributed between actions $a$ and $b$, we have:

$$EF_a - EF_b \geq 2(\gamma p_b f_b + \delta p_b f_b + \gamma \delta p_b f_b) \qquad (7)$$

Therefore,

$$\frac{EF_a}{EF_b} \geq 2(\gamma + \delta + \gamma \delta) + 1 \qquad (8)$$

This sensitivity information has been used to produce plots of the degree of accuracy required of one variable (probability or fulfilment), given the accuracy of the other variable (figure 4 and 5). In figure 4 the x-axis shows the value of the accuracy measures ($\alpha$ and $\gamma$ above) required of the probability assessments for two actions. For example, if the accuracy of the probabilities is known to $p_{a,b} \pm 0.2 p_{a,b}$ and the fulfilment is known to accuracy $f_{a,b} \pm 0.3 f_{a,b}$, then the ratio of expected fulfilments must be greater than 2.1 to guarantee the selection of the operator that returns the greatest EF. As equation 8 is symmetrical, the x-axis and labelled data lines can be used interchangeably to represent accuracy of probability or fulfilment.

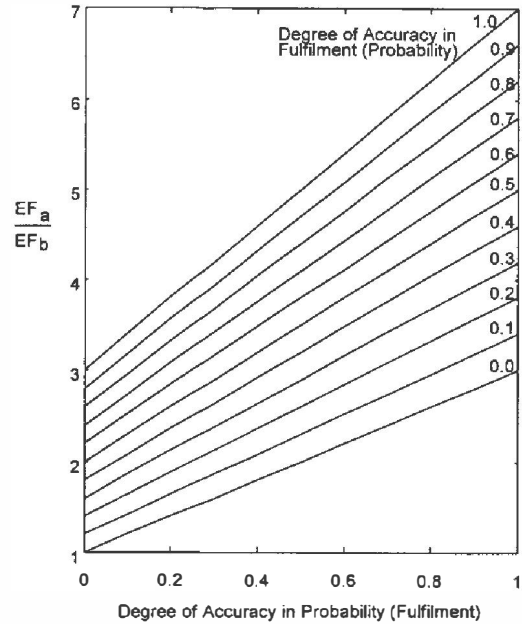

Figure 4: The degree of accuracy required of probability (or fulfilment) given the EF ratio plotted for different values of fulfilment (or probability).

Similarly, figure 5 gives a three dimensional representation of the accuracy measures for probability and fulfilment plotted against the EF ratio for two actions. Included in this plot, on the base of the graph, are the contour lines of constant EF ratio given probability and fulfilment accuracy measures. The first and last of the seven contour lines being point values at the origin and extremity of the plane.

This sensitivity data allows the knowledge engineer and user of U-Plan sufficient information to understand when the selection of competing actions of specified expected fulfilments may be inconclusive. Although no explanation process currently exists in U-Plan, this information could be utilised in an extension to U-Plan that allowed the user to question why a decision was made, or produce a confidence in specific operator selection. Both these issues are research areas in themselves, and are not examined here.

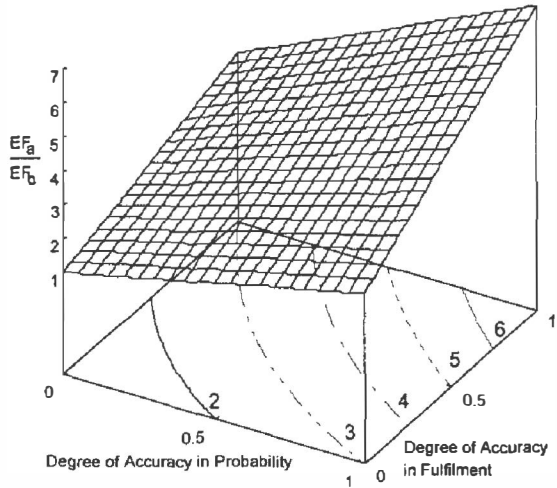

Figure 5: This figure shows the relative degree of accuracy required of probability and fulfilment measures plotted against the ratio of $EF_a$ over $EF_b$. Included on the base is the contour plot for the integer values of $EF_a/EF_b$.

## 5 PLAN REAPPLICATION

U-Plan applies plan reapplication in an attempt to determine if a plan generated for one initial P-state can be adopted for another initial P-state. The desired result being fewer plans than the number of initial P-states. This is implemented by attempting to reapply plans generated for one initial P-state to other initial P-state.

A plan is reapplicable if all the reduction operators in the plan (that are not redundant) have their preconditions met under the new initial P-state, and when applied result in the goal state being achieved. If a plan, during reapplication, fails due to the unsuccessful application of an operator, that plan is not entirely discarded. U-Plan will attempt to use the part of the plan that was successful and planning continues from the point where the plan failed. The desire is to construct plans with the same or similar strategies by reusing, at least part of, the plan at the high level of abstraction. When more than one plan partially works for a new initial P-state the best plan (Mansell 1993a) is used.

## 6 SUPER-PLANS

Once plans exists for all the P-states, with support and plausibility above some threshold, a single super-plan is constructed. This is achieved by merging the set of plans constructed for the set of initial P-states, that is applying identical operator sequences and branching at the point where plans differ. At each branch in the super-plan a knowledge acquisition operator is added, attaining the information required to select which action in the super-plan to apply next. The case may arise when the required information to differentiate between alternative branches is not available. In this case, the selection is based on the degree of evidence supporting each branch of the super-plan (see (Mansell 1993a) for more detail). The generation of a super-plan is an important attribute of U-Plan, as it presents a user with a recommended coarse of action.

## 9 CONCLUSION

U-Plan is a hierarchical planner that deals with information represented at a level of abstraction equivalent to the action being investigated. Outlined in this paper is the quantitative best-first search method employed by U-Plan for operator selection in an abstraction hierarchy. As this process is a forward propagating partial decision tree, a method for reviewing previous decisions in the light more detailed information is included. The update rules are presented in some detail, and an example of their operation presented. U-Plan has proved to be a effective planning system in the air combat domain (Mansell 1993a), and the expected fulfilment calculation a reliable formula for operator selection.

## References

W. J. Karnavas, P. J. Sanchez, and A. Terry Bahill (1993), Sensitivity Analysis of Continuous and Discrete Systems in the Time and Frequency Domains. In *IEEE Transactions on Systems, Man, and Cybernetics*, Vol. 23, No. 2, 488-501.

J. Lowrance, T. Strat, L. Wesley, T. Garvey, E. Ruspini and D. Wilkins (1991), *The Theory, Implementation, and Practice of Evidential Reasoning*. Technical report, Artificial Intelligence Centre, SRI International, California.

T. M. Mansell (1993a), A Method for planning Given Uncertain and Incomplete Information. In *Proceedings of the Ninth Conference on Uncertainty in Artificial Intelligence*, Washington DC, pg 350-358, 1993.

T. M. Mansell (1993b), Air Combat Planning Using U-Plan for Post Mission Analysis. In *Proceedings of the First Australian and New Zealand Conference on Intelligent Information Systems*, Perth, Australia, pg 644-648.

G. A. Shafer (1976), *Mathematical Theory of Evidence*. Princeton University Press, Princeton, New Jersey, USA, 1976.

D. E. Wilkins (1988), *Practical Planning: Extending the Classical AI Planning Paradigm*. Morgan Kaufmann, Los Altos, California.